\documentclass[smallextended]{spr_template/svjour3}       
\smartqed  
\usepackage[numbers]{natbib}
\usepackage[acronym,xindy,toc,nonumberlist,shortcuts]{glossaries} 
 \usepackage{floatrow}
 \usepackage{rotating}
 \usepackage[table]{xcolor}
 \usepackage{url}
 \usepackage{graphics}
 \graphicspath{{./figures/}}
 
 \usepackage{array} 
 \newcolumntype{P}[1]{>{\raggedright\arraybackslash}p{#1}}
 \newcommand{\tick}{$\surd$}
\newacronym{cpted}{CPTED}{Crime prevention through environmental design}
\newacronym{lod}{LOD}{linked open data}
\newacronym{vgi}{VGI}{volunteered geographic information}
\newacronym{hci}{HCI}{human-computer interaction}
\newacronym{gir}{GIR}{geographic information retrieval}
\newacronym{ai}{AI}{artificial intelligence}
\newacronym{ir}{IR}{information retrieval}
\newacronym{gis}{GIS}{geographic information system}
\newacronym{idf}{IDF}{Inverse Document Frequency}
\newacronym{lsa}{LSA}{Latent Semantic Analysis}
\newacronym{poipl}{POI}{points of interest}
\newacronym{poi}{POI}{point of interest}
\newacronym{vsm}{VSM}{vector space model}
\newacronym{pos}{POS}{part-of-speech}
\newacronym{gwap}{GWAP}{game with a purpose}
\newacronym{www}{WWW}{World Wide Web}
\newacronym{skos}{SKOS}{W3C Simple Knowledge Organization System}
\newacronym{rdf}{RDF}{Resource Description Framework}
\newacronym{owl}{OWL}{Web Ontology Language}
\newacronym{oaei}{OAEI}{Ontology Alignment Evaluation Initiative}
\newacronym{giscience}{GIScience}{geographic information science}
\newacronym{wsddef}{WSD}{word sense disambiguation}
\newacronym{dl}{DL}{description logic}
\newacronym{mdsmsim}{MDSM}{Matching-Distance Similarity Measure}
\newacronym{oursurvey}{GeReSiD}{Geo Relatedness and Similarity Dataset}
\newacronym{ira}{IRA}{interrater agreement}

\newacronym{irr}{IRR}{interrater reliability}
\newacronym{mds}{MDS}{multidimensional scaling}
\newacronym{lbs}{LBS}{location-based services}
\newacronym{oss}{FOSS}{free and open-source software}
\newacronym{sdts}{SDTS}{Spatial Data Transfer Standard}
\newacronym{tfidf}{TF-IDF}{Term Frequency-Inverse Document Frequency}
\newacronym{ogc}{OGC}{Open Geospatial Consortium}

\newglossaryentry{w2}{name=Web 2.0,
	description={TODO}
}

\newglossaryentry{bow}{name=bag-of-words}
\newglossaryentry{grelsim}{name=geo-semantic relatedness and similarity}

\newglossaryentry{algo}{name=\emph{Voc2WordNet}}
\newglossaryentry{algoplain}{name=Voc2WordNet}

\newglossaryentry{lodcloud}{name=LOD cloud}

\newglossaryentry{stratag}{name=Strategic Research in Advanced Geotechnologies (StratAG)}

\newglossaryentry{wn}{name=WordNet}
\newglossaryentry{osm}{name=OpenStreetMap}

\newglossaryentry{kb}{name={knowledge base}}
\newglossaryentry{gkb}{name={geo-knowledge base}}

\newglossaryentry{nuim}{name={National University of Ireland, Maynooth},
	description={TODO}
}

\newglossaryentry{ucd}{name={University College Dublin},
	description={TODO}
}

\newglossaryentry{osmsim}{name=\textsc{OSM-TagSim},
	description={TODO}
}

\newglossaryentry{lexsim}{name=\textsc{osm-sim_{lex}},
	description={TODO}
}

\newglossaryentry{simdl}{name=Sim-DL,
	description={TODO}
}

\newglossaryentry{netsim}{name=\textsc{osm-sim_{sim}},
	description={TODO}
}

\newglossaryentry{dbp}{name=DBpedia,
	description={TODO}
}

\newglossaryentry{lgd}{name=LinkedGeoData,
	description={TODO}
}

\newglossaryentry{sgw}{name=Semantic Geospatial Web,
	description={TODO}
}

\newglossaryentry{sw}{name=Semantic Web,
	description={TODO}
}

\newglossaryentry{webplatform}{name=Web platform for map personalisation and visualisation,
	description={TODO}
}

\newglossaryentry{wsd}{name=word sense disambiguation,
	description={TODO}
}

\newglossaryentry{os}{name=open source,
	description={TODO}
}

\newglossaryentry{osn}{name=OSM Semantic Network,
	description={TODO}
}

\newglossaryentry{nlp}{name=natural language processing,
	description={TODO}
}

\newglossaryentry{owc}{name=OSM Wiki Crawler,
	description={TODO}
}

\newglossaryentry{oww}{name=OSM Wiki website,
	description={\url{http://wiki.openstreetmap.org}}
}

\newglossaryentry{mdsm}{name=MDSM evaluation dataset,
	description={TODO}
}

\newcommand{\accesseddate}[0]{(acc. Apr 10, 2013)}

\newcommand{\osmtag}[1]{\emph{#1}} 

\newcommand{\footurl}[1]{\footnotemark\footnotetext{\url{#1} \accesseddate{}}}
\newcommand{\footurltwo}[2]{\footnotemark\footnotetext{\url{#1}, \url{#2}}}
\newcommand{\footurlthree}[3]{\footnotemark\footnotetext{\url{#1}, \url{#2}, \url{#3} \accesseddate{}}}
\newcommand{\urlfoot}[1]{\footurl{#1}}
\newcommand{\homepagenote}[1]{\footnote{\homepage{#1} \accesseddate{}}}
\newcommand{\homepage}[1]{\url{http://github.com/ucd-spatial/#1}}

\usepackage{graphicx}
\begin{document}

\title{An Evaluative Baseline for Geo-Semantic Relatedness and Similarity}


\author{Andrea Ballatore \and Michela Bertolotto \and David C. Wilson}


\institute{Andrea Ballatore \& Michela Bertolotto \at
              School of Computer Science and Informatics\\
              University College Dublin, Belfield, Dublin 4, Ireland\\
              \email{\{andrea.ballatore,michela.bertolotto\}@ucd.ie}           
           \and
           David C. Wilson \at
              Department of Software and Information Systems\\
				University of North Carolina, Charlotte, NC, USA\\
              \email{davils@uncc.edu}           
}

\date{Received: date / Accepted: date}

\maketitle
\begin{abstract}

In geographic information science and semantics, the computation of semantic similarity is widely recognised as key to supporting a vast number of tasks in information integration and retrieval.
By contrast, the role of geo-semantic relatedness has been largely ignored.
In \gls{nlp}, semantic relatedness is often confused with the more specific semantic similarity.
In this article, we discuss a notion of geo-semantic relatedness based on Lehrer's semantic fields, and we compare it with geo-semantic similarity. 
We then describe and validate the \emph{\gls{oursurvey}}, a new open dataset designed to evaluate computational measures of geo-semantic relatedness and similarity.
This dataset is larger than existing datasets of this kind, and includes 97 geographic terms combined into 50 term pairs rated by 203 human subjects. 
\gls{oursurvey} is available online and can be used as an evaluation baseline to determine empirically to what degree a given computational model approximates geo-semantic relatedness and similarity.

\keywords{geo-semantic relatedness \and geo-semantic similarity \and gold standards \and geo-semantics \and cognitive plausibility \and GeReSiD}
\end{abstract}

\section{Introduction}
\label{sec:intro}

Is \emph{lake} related to \emph{river}?
Is \emph{road} related to \emph{transportation}?
Are \emph{mountain} and \emph{hill} more related than \emph{mountain} and \emph{lake}?
While it may seem natural to answer yes to all of these questions, the logical and computational formalisation of why this is the case has raised considerable interest in philosophy, psychology, linguistics and, more recently, in computer science.
The human ability to detect semantic relatedness is essential to perform key operations in communication, such as word-sense disambiguation (e.g. interpreting \emph{bank} as financial institution or as the terrain alongside the bed of a river), reducing semantic ambiguity and increasing efficiency in meaning-creation and sharing.
The human cognitive apparatus possesses a remarkable ability to detect co-occurrence patterns that are not due to chance, but that indicate the existence of some semantic relation between the terms.

\emph{Semantic similarity} has been identified as a particular subset of this general notion of semantic relatedness.
While semantically \emph{related} terms are connected by any kind of relation, semantically \emph{similar} terms are related by synonymy, hyponymy, and hypernymy, all of which involve an \emph{is a} relation.
In this sense, \emph{train} and \emph{bus} are intuitively similar (they are both means of transport), whilst \emph{bus} and \emph{road} are related but not similar (i.e. they often co-occur but with different roles).
Semantic similarity relies on the general cognitive ability to detect similar patterns in stimuli, which attracts considerable attention in cognitive science.
Notably, \citet{goldstone:2005:similarity} stated that ``assessments of similarity are fundamental to cognition because similarities in the world are revealing. The world is an orderly enough place that similar objects and events tend to behave similarly'' (p. 13).
Therefore, the vast applicability of semantic similarity in computer and information science should come as no surprise.

In \gls{giscience}, the theoretical and practical importance of geo-semantic similarity has been fully acknowledged, resulting in a growing body of research \cite{bakillah:2009:simnet,Ballatore:2013:lexicaldefsim,janowicz:2011:semantics}. 
By contrast, the importance of semantic relatedness, which is widely studied in the non-geographic domain, has been almost completely ignored, with the exception of the works by \citet{hecht:2008:geosr} and \citet{hecht:2012:explanatorysr}.
Computational measures of semantic relatedness play a pivotal role in \gls{nlp}, \gls{ir}, and \gls{wsd}, providing access to deeper semantic connections between words and sets of words.
Despite the large number of existing measures, their rigorous evaluation still constitutes an important research challenge \cite{ferrara:2013:evalrelatedness}.
 
This article contributes to \gls{giscience} and semantics in the following ways.
First, we discuss in detail the notion of geo-semantic relatedness, drawing on Lehrer's theory of semantic fields, which consist of sets of terms covering a restricted semantic domain.
Geo-semantic relatedness is defined with respect to specifiable geographic relations between terms, and is compared and contrasted with the more widely studied geo-semantic similarity.
\glsreset{oursurvey}
Second, we have developed and validated the \gls{oursurvey}, tackling the complex issue of the evaluation of computational measures of \gls{grelsim}.
In this new dataset, we have collected psychological judgements about 50 pairs of terms, covering 97 unique geographic terms, from 203 human subjects. 
The human judgements in \gls{oursurvey} focus explicitly on geo-semantic relatedness and similarity between geographic terms.

The resulting dataset provides an evaluation test bed for geo-semantic relatedness and similarity.
This is compared against the existing human-generated gold standards used to assess computational measures of semantic relatedness and similarity, highlighting the limitations of such datasets.
Such an evaluative baseline constitutes a valuable ground truth against which computational measures can be assessed, providing empirical evidence about the cognitive plausibility of the measures.
\gls{oursurvey} can inform research in geo-semantics, indicating to what degree computational approaches match human judgements.   
More specifically the contribution of this evaluative baseline consists of the following aspects:

\begin{itemize}
  \item \gls{oursurvey} covers a sample of geographic terms larger than existing similarity datasets, including 97 natural and man-made unique terms, grouped in 50 unique pairs. 
  Psychological judgements of geo-semantic relatedness and similarity were collected separately on the 50 pairs. 
  \item \gls{oursurvey} includes a sample of evenly distributed relatedness/similarity judgements, ranging from near-synonymity to no relationship between the terms.
  Our methodology is described explicitly and precisely, in order to provide practical guidelines to construct similar datasets.
  \item Unlike existing datasets, the semantic judgements on the term pairs contained in \gls{oursurvey} are analysed with respect to \gls{ira} and \gls{irr}.
  \item The psychological judgements in \gls{oursurvey} can be observed as the mean of relatedness/similarity of the pairs, using correlation coefficients of relatedness/similarity rankings (such as Spearman's $\rho$ or Kendall's $\tau$).
  Alternatively, the data can be interpreted as categorical, using Cohen's kappa or Fisher's exact test \cite{banerjee:1999:beyond} to evaluate the computational measure. 
  \item \gls{oursurvey} is an open dataset freely available online.\homepagenote{Datasets} Both raw data and the resulting dataset are available.
\end{itemize}

\noindent The remainder of this article is organised as follows.
Section \ref{sec:simVsRel} discusses in depth the two key notions of geo-semantic relatedness and similarity, proposing a synthetic definition.
Section \ref{sec:relwork} summarises existing datasets for semantic relatedness and similarity, with particular attention to those restricted to the  geographic domain.
The new evaluative baseline, \gls{oursurvey}, is outlined, analysed and discussed in Section \ref{sec:geresid}.
Conclusions and directions for future research are indicated in Section \ref{sec:concl}.

\section{Geo-semantic relatedness and similarity}
\label{sec:simVsRel}



%


This section introduces the notion of \emph{geo-semantic relatedness}, comparing it and contrasting it with \emph{geo-semantic similarity}.
In the \gls{nlp} literature, several terms are used inconsistently, including semantic relatedness, relational similarity, taxonomical similarity, semantic association, analogy, and attributional similarity \citep{turney:2006:similarity}.
These terms are often used interchangeably \citep{budanitsky:2006:evaluating}.
A striking example of this tendency is the article title `WordNet::Similarity: Measuring the relatedness of terms' \citep{pedersen:2004:wordnetsim}.

In natural language, terms are connected by an open set of semantic relations.
Common semantic relations are synonymy ($A$ coincides with $B$), antonymy ($A$ is the opposite of $B$), hyponymy ($A$ is a $B$), hypernymy ($B$ is a $A$), holonymy ($A$ is whole of $B$), meronymy ($A$ is part of $B$), causality ($A$ causes $B$), temporal contiguity ($A$ occurs at the same time as $B$), and function ($A$ is used to perform $B$).
\citet{khoo:2006:semantic} have surveyed these semantic relations, whilst \citet{morris:2004:nonclassical} have explored other non-classical semantic relations.
As \citet{khoo:2006:semantic} remarked, semantic relations are characterised by productivity (new relations can be easily created), uncountability (semantic relations are an open class and cannot be counted), and predictability (they follow general, recurring patterns). 
In the geographic domain, spatial relations such as proximity ($A$ is near $B$), and containment ($A$ is within $B$) have an impact on semantics \citep{schwering:2005:spatial}.

Before providing our definition of geo-semantic relatedness and similarity, it is beneficial to review the semantics of these terms in the literature.
In the context of semantic networks, \citet{rada:1989:development} suggested that semantic relatedness is ``based on an aggregate of the interconnections between the terms'' (p. 18).
To obtain semantic similarity, the observation must be restricted to taxonomic \emph{is\_a} relationships between terms.
\citet{resnik:1995:using} followed this approach, and defined semantic similarity and relatedness as follows: 
``Semantic similarity represents a special case of semantic relatedness: for example, cars and gasoline would seem to be more closely related than, say, cars and bicycles, but the latter pair are certainly more similar'' (p. 448). 

More recently, \citet{turney:2006:similarity} added a further distinction between `attributional' and `relational similarity.'
Following the approach outlined by \citet{medin:1990:similarity}, `attributes' are statements about a term that take only one parameter, e.g. $X~is~red$, $X~is~long$.
Therefore, attributional similarity measures the correspondence between the attributes of the two terms.
`Relations,' on the other hand, are statements that take two or more parameters, e.g. $X~is~a~Y, X~is~longer~than~Y$.
Hence, relational similarity is based on the common relations between two pairs of terms \citep{turney:2006:similarity}.
On these assumptions, synonymy is seen as a high degree of attributional similarity between two terms, e.g. $<$river,stream$>$.
Analogy, by contrast, is characterised as a high degree of relational similarity between two pairs of terms, e.g. $<$boat,river$>$ and $<$car,road$>$.
The next sections discuss \gls{grelsim} in detail.

\subsection{Geo-semantic relatedness}

A general notion of \emph{relatedness} in the geographic context was stated in Tobler's first law, which asserts that everything is related to everything else, but near things are more related than distant things \citep{tobler:1970:computer}.
While this law was formulated to express intuitively the high spatial autocorrelation of many geographic phenomena, it has generated several responses in \gls{giscience}.
For example, in the context of information visualisation, \citet{montello:2003:firstlawcognitive} have proposed the \emph{first law of cognitive geography}, which states that ``people believe closer things to be more similar than distant things'' (p. 317).
Applying the same intuition to the domain of geo-semantics, we assert that two terms are \emph{geo-semantically related} to the degree to which they refer to entities or phenomena connected via specifiable relations grounded in the geographic dimension.

To define a notion of geo-semantic relatedness, we rely on the notion of \emph{semantic field}.
According to \citet{lehrer:1985:influence}, a semantic field is ``a set of lexemes which cover a certain conceptual domain and which bear certain specifiable relation to one another'' (p. 283).
While a `domain' is an epistemological notion referring to a subset of human knowledge and experience (e.g. geography, politics, medicine, etc.), a semantic field is a more specific linguistic notion that refers to a set of lexemes utilised to describe a domain.
For example, a semantic field might be formed by terms \emph{train}, \emph{bus}, \emph{trip}, \emph{fare}, \emph{delay}, \emph{accident}, etc., which are all connected to the underlying term of transportation, and commonly used to generate observations on the domain of mobility.

Terms appear to be semantically related to the degree to which they belong to the same semantic field, and can indeed belong to different semantic fields.
Semantic fields are neither static nor well-defined sets, but rather fuzzy configurations that shift over time, and across different agents and information communities.
The condition of \emph{specifiability} of relations emphasises the fact that random co-occurrence has no impact on semantic relatedness.
If a relation is not specifiable, the co-occurrence of the two terms must be random.
A term has a certain degree of \emph{centrality} in a semantic field, i.e. the density of connectedness with other terms.
For example, in the aforementioned semantic field on transportation, \emph{car} is more central than \emph{delay}.
Similarly, in a semantic field on social life, \emph{car} is likely to be less central than \emph{restaurant} or \emph{pub}.

Geo-semantic relatedness can therefore be defined as a specific sub-domain of semantic relatedness, focusing on relations grounded in the geographic dimension, i.e. relations in which at least one of the terms has a spatial dimension.
Examples of geo-semantically related terms are \emph{judge}, \emph{trial}, and \emph{tribunal}, where \emph{tribunal} has a strong geographic component that grounds the other terms geographically.
A computational measure of geo-semantic relatedness has to aggregate and quantify the intensity of such relations between two terms, providing a useful tool for several complex tasks.
For example, terms \emph{river} and \emph{flood} should be more geo-semantically related than \emph{vehicle} and \emph{car}, which possess a less prominent geographic component.
Acknowledging the fact that most terms in natural language have some degree of geographic ground, we express this approach to geo-semantic relatedness following Tobler's first law of geography:
\begin{quote}
\emph{Every term is geo-semantically related to all other terms, but terms that co-occur with specifiable geographic relations are more related than other terms.}
 \end{quote}

\noindent In other words, every term can in principle have some degree of geo-semantic relatedness to any other term, but terms that co-occur in observations bearing specifiable relations tend to be more geo-semantically related than those that do not.
This formulation puts terms in relation to human spatial experience from which terms arise, suggesting indistinct, gradual, and shifting boundaries between geo-related and unrelated terms.

In this sense, geo-semantic relatedness is intrinsically fuzzy, admitting a continuous spectrum of relatedness rather than a binary classification (i.e. \emph{related} or \emph{unrelated}).
Highly related terms belong to the same semantic field.
The same terms can belong to several overlapping semantic fields.
Relatedness involves all semantic relations, including synonymy, antonymy, hyponymy, hypernymy, holonymy, meronymy, causality, temporal contiguity, function, proximity, and containment.
This law applies both to natural language, where geographic terms can be highly imprecise and vague, and to scientific conceptualisations, which generally aim at stricter semantics.

Surprisingly, in \gls{giscience} semantic relatedness has been almost completely ignored, with two notable exceptions \cite{hecht:2008:geosr,hecht:2012:explanatorysr}.
In order to explore semantically and spatially related entities in Wikipedia, \citet{hecht:2008:geosr} developed ExploSR, a graph-based relatedness measure.
ExploSR computes a semantic relatedness score of two articles by assigning weights to spatially-referenced articles in the Wikipedia Article Graph.
More recently, the \emph{Atlasify} system generates human-readable explanations of the relationship between terms to support exploratory search \cite{hecht:2012:explanatorysr}.

Geo-semantic relatedness can be informed by ideas developed in the area of text mining.
The latent Dirichlet allocation (LDA) adopts a probabilistic approach to cluster highly semantically-related terms in a text corpus \cite{blei:2003:lda}.
LDA was extended to include a geographic dimension into the Location Aware Topic Model (LATM) \cite{wang:2007:mining}.
LATM quantifies the geo-semantic relatedness between keywords, topics, and geographic locations, adopting a fully distributional approach.


\subsection{Geo-semantic similarity}

While geo-semantic relatedness of terms can be based on co-occurrence in observations, geo-semantic similarity of terms can only be determined through the analysis of the terms' attributes and relations.
Geo-semantic similarity is a subset of geo-semantic relatedness: all similar terms are also related, but related terms are not necessarily similar.
The relations considered for geo-semantic similarity include only synonymy, hyponymy, and hypernymy.
Unlike geo-semantic relatedness, geo-semantic similarity has been deeply explored by the \gls{giscience} community, and is recognised as one of the key concepts of geo-semantics \cite{kuhn:2013:cognling}.

Several theories of similarity have been used to conceptualise and measure geo-semantic similarity, including featural, transformational, geometric, and alignment models \cite{schwering:2008:approaches,janowicz:2008:semantic,janowicz:2011:semantics,schwering:2009:hybsimgir}.
Specific techniques have been devised for specific knowledge-representation formalisms \cite{janowicz:2007:algorithm,rodriguez:2004:comparing}.
More recently, graph-based \cite{Ballatore:2012:geographic} and lexical techniques \cite{Ballatore:2012:jury,Ballatore:2013:lexicaldefsim} have been investigated in the emerging area of \gls{vgi}.
These works tend to focus on the conceptual level, computing the similarity of abstract geographic terms (e.g. \emph{city} and \emph{river}), rather than the instance level (e.g. \emph{New York} and \emph{Danube}).


Beyond the specificities of such approaches, we can state that terms $A$ and $B$ are semantically similar with respect to $C$, where $C$ is a set of attributes and relations, also known as \emph{context} \cite{kessler:2007:similarity}.
The context $C$ focuses on the typical spatial organisation and appearance of the entity identified by the term (e.g. shape, size, material composition).
Alternatively, the similarity of $A$ and $B$ can be measured with respect to their \emph{affordances}, i.e. the possibilities that an entity offers to humans \citep{janowicz:2007:affordance}.

As observed in relation to geo-semantic relatedness, all terms can be geo-semantically similar to some limited extent, and geo-semantic similarity is therefore best modelled as a continuous spectrum, rather than a binary classification.
For example, terms \emph{restaurant} and \emph{continent} are similar with respect to the fact that they both refer to geographically-grounded entities.
To capture this idea at the linguistic level that is relevant to this discussion, we adopt the approach outlined in \cite{Ballatore:2013:lexicaldefsim}.
Considering the terms used in lexical definitions of terms, we state recursively that:
\begin{quote}
\emph{All terms are geo-semantically similar, but geographic terms described using the same terms are more similar than other terms.}
\end{quote}

\noindent A geo-semantic similarity measure has to quantify the similarity of two terms into a score, enabling a number of semantic tasks in \gls{ir} and information integration.
For example, terms \emph{restaurant} and \emph{pub} are very similar because they share similar spatial organisation and affordances.
\emph{Houses} and \emph{schools} are geo-semantically similar with respect to their spatial organisation of parts and can be described as having walls, windows, doors, a roof, etc.
\emph{Roads} and \emph{rivers} show similar affordances -- they can be used for transportation.

\section{Semantic relatedness and similarity gold standards}
\label{sec:relwork}

\label{sec:relwork_geoSimilarity}
\label{sec:relwork_goldStandards}

Semantic similarity and relatedness measures can be evaluated against a human-generated set of psychological judgements.
This section gives an overview of published similarity and relatedness gold standards, mostly from psychology and computational linguistics.
The term `gold standard' 
is described by the Oxford Dictionary of English as ``a thing of superior quality which serves as a point of reference against which other things of its type may be compared.''\urlfoot{http://oxforddictionaries.com/definition/gold+standard}
In computer science, the term is used to describe high-quality, human-generated datasets, capturing human behaviour in relation to a well-defined task.
Such datasets can then be used to assess the performance of automatic approaches, by quantifying the correlation between the machine and the human-generated data.

\subsection{Cognitive plausibility}

In a seminal discussion on expert systems, \citet{strube:1992:role} argued that knowledge engineering should strive towards increasing the \emph{cognitive adequacy} of computational systems, defined as their `degree of nearness to human cognition' (p. 165).
In the context of \gls{giscience}, geo-relatedness or geo-similarity measures need not replicate the workings of human mind in their entirety (defined as \emph{absolutely strong adequacy}), but should aim at what Strube called \emph{relatively strong adequacy}, i.e. the ability of the system to function like a human expert in a circumscribed domain.
Following this approach, we adopt the notion of \emph{cognitive plausibility} to assess to what degree a measure mimics human behaviour \cite{kessler:2011:difference}.

In order to quantify the cognitive plausibility of a computational semantic relatedness or similarity measure, two complementary approaches can be adopted: (1) psychological evaluations, and (2) task-based evaluations.
In psychological evaluations, human subjects are asked to rank or rate term pairs.
These rankings or ratings are then compared with computer-generated rankings, usually using correlation as an indicator of performance.
Alternatively, human subjects can perform a task based on the assessment of relatedness or similarity, such as \gls{wsd}, and the cognitive plausibility of the measure is observed indirectly in the results of the task, using for example precision and recall measures.
Such human-generated datasets are used as gold standards.

The usage of gold standards is common in \gls{nlp} tasks, such as part-of-speech tagging, entity resolution, and \gls{wsd} \citep{schutze:1998:automatic,toutanova:2003:feature,cimiano:2005:towards,pedersen:2009:wsd}. 
Adopting this approach, a technique or a model can be deemed to be more or less plausible by observing its correlation with human-generated results.  
Such datasets are created by combining the results from a number of human subjects who perform a given task, either under controlled conditions, or through online forms.
To be considered valid by a research community, a gold standard needs to meet certain criteria, such as coverage, quality, precision, and inter-subject agreement.
Disagreements about the validity of a gold standard are quite common and, when weaknesses are uncovered, a gold standard can be demoted to a golden calf \cite[e.g.][]{kaptchuk:2001:double}.

The intrinsic high subjectivity of relatedness and similarity rankings makes the collection and validation of gold standards complex and challenging.
Although task-based evaluations might appear more `objective,' they are equally affected by subjectivity: ultimately, relatedness-based or similarity-based tasks are generated, interpreted, and validated by human subjects.
Acknowledging the unlikelihood of total agreement, the reliability of a similarity evaluation should be grounded in stability over time, consistency across different datasets, and reproducibility of psychological results.
Ideally, both evaluation approaches should show convergent, cross-validating results: a strong correlation is expected between the cognitive plausibility of a measure and its performance in similarity-based tasks.


\subsection{Comparison of relatedness and similarity gold standards}

Over the past 50 years, several authors investigating semantic issues in psychology, linguistics, and computer science created datasets focused on semantic similarity and, more recently, semantic relatedness.
The first similarity gold standard was published in 1965, in a article in which \citet{rubenstein:1965:contextual} collected a set of 65 word pairs ranked by their synonymy.
Following a similar line of research, \citet{miller:1991:contextual} published a similar dataset with 30 word pairs in 1991.
More recently, \citet{finkelstein:2002:placing} created the WordSimilarity-353 dataset, which contains 353 word pairs actually ranked by semantic relatedness.\footurl{http://www.cs.technion.ac.il/~gabr/resources/data/wordsim353}
The dataset was subsequently extended to distinguish between similarity and relatedness \citep{agirre:2009:study}.\footurl{http://alfonseca.org/eng/research/wordsim353.html}
In a study of the retrieval mechanism of memories, \citet{nelson:2005:what} collected associative similarity ratings for 1,016 word pairs.

\begin{sidewaystable}
\vspace*{12.5cm} 
\footnotesize
\rowcolors{2}{gray!15}{white}
\begin{tabular}{P{8em}P{14em}P{17em}P{2em}P{2em}P{2em}}
\hline
Reference & Subjects & Terms \& term pairs & \textsc{rel} & \textsc{sim} & \textsc{geo} \\
\hline 

\citet{rubenstein:1965:contextual}
& 51 paid college undergrads: group I (15 subjects), group II (36 subjects). 
& 48 terms (ordinary English words); 65 term pairs, ranging from highly synonymous pairs to semantically unrelated pairs.
& 
& \tick 
&  
\\
\citet{miller:1991:contextual}
& 38 undergraduate students. US English native speakers. 
& 40 terms selected from \citet{rubenstein:1965:contextual}; 32 term pairs. 
& 
& \tick 
&  
\\ 

\citet{finkelstein:2002:placing,agirre:2009:study} 
& 13 experts for first set (153 pairs), 16 experts for second set (200 pairs). Near-native English proficiency. 
& 346 terms (manually selected nouns and compound nouns); 353 term pairs.  
& \tick 
& \tick 
&  
\\    
\citet{rodriguez:2004:comparing} 
& 72 paid undergrad students (two groups of 36 people). US English native speakers. 
& 33 geographic terms from WordNet and SDTS; 10 sets of 10 or 11 term pairs. 
& 
& \tick 
& \tick 
\\
  
\citet{nelson:2005:what}
& 94 undergraduate students rewarded with academic credits 
& 1,016 term pairs selected unsystematically from a cued recall database of 2,000+ pairs. 
& \tick 
&  
&  
\\

\citet{janowicz:2008:study}
& 28 unpaid subjects (20-30 years of age).
& Six geographic terms related to bodies of water.
& 
& \tick 
& \tick 
\\ 
\textbf{\gls{oursurvey}} (see Section \ref{sec:geresid})
& 203 unpaid English native speakers.
& 97 geographic terms from \gls{osm}; 50 term pairs.
& \tick 
& \tick 
& \tick 
\\  
\hline  
\end{tabular}
\caption{Semantic relatedness and similarity gold standards}
\label{table:simGoldStandards}
\end{sidewaystable}


A smaller number of geo-semantic similarity datasets have been generated in the areas of \gls{giscience} and \gls{gir}.
In this area, \citet{janowicz:2008:study} conducted a study on the cognitive plausibility of their \gls{simdl} similarity measure.
However, the study was conducted in German on a very small set of terms, and for this reason it is difficult to reuse in different contexts. 
In order to evaluate their \gls{mdsmsim}, \citet{rodriguez:2004:comparing} collected similarity judgements about geographic terms, including large natural entities (e.g. \emph{mountain} and \emph{forest}), and man-made features (e.g. \emph{bridge} and \emph{house}).
Before \gls{oursurvey}, the \gls{mdsm} was the largest similarity gold standard for geographic terms.
For this reason, this dataset was utilised to carry out the evaluation of network-based similarity measures \citep{Ballatore:2012:geographic}.
In contrast, geo-semantic relatedness has been largely ignored in the geospatial domain.  


The salient characteristics of these gold standards are summarised in Table \ref{table:simGoldStandards}, detailing their human subjects, the terms and term pairs.
For each dataset, the table shows whether they focus on semantic relatedness (\textsc{rel}), semantic similarity (\textsc{sim}), and exclusively on the geographic domain (\textsc{geo}).
The existing datasets are compared with \gls{oursurvey}, the gold standard described in Section \ref{sec:geresid}, and have several limitations.  
First, the procedure followed to construct the datasets is usually only sketched and not described in detail.
Second, the size of the datasets tends to be rather small.

The size of such datasets can be observed along three dimensions: number of human subjects, number of terms, and number of term pairs.
A clear trade-off exists between number of human subjects and number of term pairs. 
Furthermore, most datasets do not capture the distinction between semantic similarity and relatedness, and do not analyse the \gls{ira} and \gls{irr}. 
It is important to note that most authors did not have the explicit intention to construct gold standards, but rather to analyse specific aspects of semantic similarity or relatedness.
However, in some cases, these datasets have been treated as gold standards in the subsequent literature \citep{rubenstein:1965:contextual,miller:1991:contextual}.
To the best of our knowledge, only WordSimilarity-353 was explicitly designed to be a generic gold standard. 

Some of these gold standards have been extensively utilised to assess general term-to-term similarity measures \citep{rubenstein:1965:contextual,miller:1991:contextual,finkelstein:2002:placing}.
In the geographic context, only the \gls{mdsm} is suitable to evaluate semantic similarity of geographic terms \citep{rodriguez:2004:comparing}.
However, no existing dataset focusing on geographic terms accounts explicitly for the difference between semantic relatedness and semantic similarity.


\section{Geo Relatedness and Similarity Dataset (GeReSiD)}
\label{sec:geresid}
\glsreset{oursurvey}

This section presents the \gls{oursurvey}, a dataset of human judgements that we have developed to provide a ground truth for the assessment of computational relatedness and similarity measurements.
\gls{oursurvey} captures explicitly the difference between \gls{grelsim} on a sample of geographic terms larger than existing similarity datasets surveyed in Section \ref{sec:relwork_goldStandards}, including both natural and man-made terms. 
In order to ensure its validity as a gold standard, it focuses on a sample of evenly distributed relatedness/similarity judgements, ranging from very high to very low.
Section \ref{sec:eval_surveyDesign} describes our methodology precisely, in order to provide guidelines on constructing datasets to ground the evaluation of measures of \gls{grelsim}.
Subsequently, Section \ref{sec:eval_surveyResults} outlines the results obtained from the online survey.

\subsection{Survey design}
\label{sec:eval_surveyDesign}

\glsreset{oursurvey}
The psychological judgements about \gls{grelsim} were collected via an online survey, through an interactive Web interface specifically designed for this purpose.
Online surveys constitute a powerful research tool, with well-known advantages and disadvantages \cite{Wright:2005:researching}.
Given the focus of this study on generic terms found in web maps, subjects involved in projects such as \gls{osm} represent an ideal virtual community of map users and producers to conduct a psychological evaluation.
An online survey is an inexpensive and effective way to reach these online communities.

A cross-disciplinary consensus exists on the fact that semantic judgements are affected by the \emph{context} in which the terms are considered \citep{rodriguez:2004:comparing,janowicz:2008:semantic}.
\citet{rodriguez:2004:comparing} asked their subjects to rank geographic terms in the following contexts: `null context,' `play a sport,' `compare constructions,' and `compare transportation systems.'
The subjects' attention was therefore focused on specific aspects of the terms being analysed, rather than on the terms in an unspecified setting.

Although context affects the assessment of semantic similarity, in this survey we aim at capturing the overall difference between semantic relatedness and similarity of terms, without focusing on specific aspects of the conceptualisation.
This comparison is an important research topic, frequently mentioned but rarely addressed directly through empirical evaluation. 
Introducing specific contexts into our survey would increase the complexity of the study by introducing new biases, making the direct comparison between similarity and relatedness problematic.
For example, adding a specific context does not increase the inter-subject agreement: in their evaluation, \citet{rodriguez:2004:comparing} report a considerably lower association between subjects in the case of context-specific questions (mean Kendall's W being $.5$), than with a-contextual questions (mean W = $.68$).
Moreover, specific contexts would introduce specific biases, which are beyond the scope of \gls{oursurvey}. 

As a solution to these issues, we frame the evaluation in the general context of popular \emph{web maps}, in which geographic terms are most frequently visualised and utilised by users.
This way, the subjects are induced to use their own conceptualisation of the geographic entities. 
As happens with semantic judgements, subjectivity inevitably affects the subjects' choices.
In this study, subjects are free to choose what properties they consider most relevant to the comparison, and the mean of their ratings quantifies the perceived inter-subject similarity and relatedness of the terms.
While the study of the context is beyond the scope of this survey, it certainly represents an important direction for future work.

The geographic terms included in this survey are taken from the \gls{osm} project.
In our previous work, we extracted the lexicon utilised in \gls{osm} into a machine-readable vocabulary, the \gls{osn} \cite{Ballatore:2012:survey}.
To date, the \gls{osn} contains a total of about 4,300 distinct terms, called `tags' in the project's terminology.
From this large set of geographic terms, a suitable sample had to be selected.
To be included, a term had to be clearly intelligible, well defined on the \gls{osn}, as culturally-unspecific as possible, and present in the actual \gls{osm} vector map.
Following these criteria, we manually selected a set $C$ of 400 terms, including a wide range of natural and man-made entities, such as `sea,' `lighthouse,' `landfill,' `valley,' and `glacier.'
Using the terms in $C$, we defined a set $P$ containing all possible pairs of geographic terms $\langle a,b \rangle$ where $a,b \in C$, for a total of 160,000 pairs.
We subsequently removed from $P$ symmetric pairs (e.g. removing $\langle b,a \rangle$ when $\langle a,b \rangle$ is defined) and identities (e.g. $\langle a,a \rangle$), resulting in 76,000 valid pairs.   

In order to detect issues in the survey, a pilot study was then conducted with 12 graduate students at University College Dublin.
A set $P_{rand}$ was constructed by selecting 100 pairs randomly from $P$.
Each pair was associated with a 5-point Likert scale, ranging from low to high relatedness/similarity. 
The subjects were asked to rate each pair both for semantic relatedness and similarity, and were then interviewed informally, to obtain direct feedback about the survey.
Several useful observations were obtained from this pilot survey. 
First, most subjects found the test too long.
A smaller sample size had to be selected, considering a trade-off between number of pairs and the completion time, in order to ensure that enough subjects would complete the task without losing concentration.
Based on the opinion of subjects, we identified 50 pairs as the maximum size of the task, with a completion time of around five minutes, suitable for an unpaid online questionnaire.

In the \gls{osm} semantic model, tags are made of a key and a value (e.g. \osmtag{amenity=school}).
In the pilot survey, this formalism had to be explained to the subjects, who  generally found it confusing.
For example, the psychological comparison between \osmtag{amenity=school} and \osmtag{amenity=community\_centre} was influenced by the shared word `amenity,' which is highly generic and ambiguous.
To make the dataset independent from the peculiar \gls{osm} tag structure, we extracted short labels for all the 400 terms from the terms' definitions.
For example, \osmtag{amenity=food\_court} was labelled as `food court,' \osmtag{shop=music} as `music shop.'
In order to increase their semantic clarity, the terms were manually mapped to the corresponding terms in \gls{wn} (see Table \ref{table:geresid_terms}).

The fully random set of 100 pairs $P_{rand}$ used in the pilot survey obtained a distribution heavily skewed towards low similarity and relatedness.
To reach a more uniform distribution, we introduced a partial manual selection in the process.  
In order to obtain an even distribution in the resulting relatedness and similarity scores, we manually extracted from the pilot survey a set of 50 pairs rated by the 12 subjects as highly related/similar pairs ($P_{high}$), and 50 middle relatedness/similarity pairs ($P_{med}$).
It is worth noting that while the selection of highly related/similar pairs is intuitive, middle-relatedness/similarity pairs is more challenging, and requires dealing with highly subjective conceptualisations.
This aspect is reflected in the survey results (see Section \ref{sec:eval_surveyResults}).
The final set of 50 pairs for the questionnaire $P_{q}$ was assembled from the following elements:  
\begin{itemize}
  \item 16 high-relatedness/similarity pairs (random sample from $P_{high}$)
  \item 18 middle-relatedness/similarity pairs (random sample from $P_{med}$)
  \item 16 low-relatedness/similarity pairs (random sample from $P$)
\end{itemize}

\begin{table}[t] 
\footnotesize
\begin{tabular}{lll}
\hline
Term & \gls{osm} tag & \gls{wn} synset \\
\hline
 bay & natural=bay & bay\#n\#1\\ 
 sea & place=sea & sea\#n\#1\\ 
basketball court & sport=basketball & basketball court\#n\#1\\ 
beauty parlor & shop=beauty & beauty\_parlor\#n\#1\\ 
floodplain & natural=floodplain & floodplain\#n\#1\\ 
greengrocer & shop=greengrocer & greengrocer\#n\#1\\ 
historic castle & historic=castle & castle\#n\#2\\ 
motel & tourism=motel & motel\#n\#1\\ 
political boundary & boundary=political & boundary\#n\#1\\ 
school & amenity=school & school\#n\#1\\ 
stadium & building=stadium & stadium\#n\#1\\ 
\ldots & \ldots & \ldots \\
\hline
 \end{tabular} 
\caption{Sample of terms in \protect\gls{oursurvey}. The dataset contains 97 geographic terms.}
\label{table:geresid_terms}
\end{table}

\noindent The pilot survey also showed clearly that assigning both the relatedness \emph{and} the similarity tasks to the same subject was impractical, and was deemed confusing by all subjects who did not possess specific expertise in linguistics.
For this reason, we opted to assign randomly only one task to each subject, either on relatedness or similarity, without trying to explain to them the technicalities of this distinction.
Instead, we relied on the subjects' inductive understanding of the task through correct examples.
Thus, in order to collect reliable judgements on similarity and relatedness, we defined two separate questionnaires, one on relatedness ($Q_{REL}$), and one on similarity ($Q_{SIM}$).
The two questionnaires were identical, with the exception of the description of the task, and the labels on the Likert scale (one with a `dissimilar-similar' scale, the other with `unrelated-related').

To avoid terminological confusion, the survey was named `Survey on comparison of geographic terms,' without mentioning either `similarity' or `relatedness' in the introductory text.
The examples used to illustrate semantic relatedness (\emph{apples - bananas}, \emph{doctor - hospital}, \emph{tree - shade}) and similarity (\emph{apples - bananas}, \emph{doctor - surgeon}, \emph{car - motorcycle}) were based on those by \citet{mohammad:2012:distributional}.
A random redirection to either $Q_{REL}$ or $Q_{SIM}$ was then implemented to ensure the random sampling of subjects into two groups, one for similarity and one for relatedness.
As the similarity judgement was reported as more difficult than relatedness, we set the probability of a random redirection to $Q_{SIM}$ at $p = .65$, to obtain more responders for similarity.
Each subject was only exposed to one of the two questionnaires.

Six general demographic questions about the subject were included: age group, mother tongue, gender, and continent of origin.
A textbox was available to type feedback and comments about the survey.
The core of each questionnaire was the seventh question, i.e. the relatedness or similarity rating task.
The subject had to rate 50 pairs of geographic terms based on their relatedness or similarity, on a 1 to 5 Likert scale.
Although the impact of size of the Likert scale, typical options being 5, 7 or 10, is debated in the social sciences, it has little impact on the rating means \citep{dawes:2008:data}.
If the terms were not clear to the user, a `no answer' option had to be selected.

Another aspect discussed in the similarity psychological literature is the counterintuitive fact that similarity judgements tend to be asymmetric (e.g. $sim(building,hospital) \neq sim(hospital,building)$) \citep{tversky:1977:features}.
As this aspect is outside the scope of this study, the order in each pair $\langle a,b \rangle$ was randomised to limit the symmetric bias, i.e. the potential difference between $sim(a,b)$ and $sim(b,a)$ from the subject's perspective.
Moreover, a fixed presentation order of pairs can trigger specific semantic associations between terms, and would reduce the quality of the last pairs, rated when the subjects are more likely to be tired.
To reduce this sequential-ordering bias, the presentation order of the pairs was randomised automatically for each subject at the Web interface level.


At the end of the design process, the survey dataset contained 50 pairs of geographic terms to be rated on 5-point Likert scales, including 97 \gls{osm} terms, with three terms being repeated twice.
The pairs were selected to ensure an even distribution between low, medium and high relatedness/similarity.
The rating was to be executed in two independent questionnaires, one for semantic similarity ($Q_{SIM}$) and one for semantic relatedness ($Q_{REL}$), randomly assigned to the human subjects.
In February 2012, the survey was disseminated in \gls{osm} and \gls{gis}-related forums and mailing lists.

\subsection{Survey results}
\label{sec:eval_surveyResults}
\glsreset{oursurvey}
 
The online questionnaires on relatedness and similarity received 305 responses, 124 for relatedness and 181 for similarity.
Given the nature of online surveys, particular attention has to be paid to the agreement between the human subjects, and the detection of unreliable and random answers. 
\glsreset{irr}
\glsreset{ira} 
In this survey, raters expressed quantitative judgements on \gls{grelsim} on a 5-point Likert scale.
Two important statistical aspects to be discussed are the \gls{irr} and the \gls{ira} \citep{lebreton:2008:answers}.
\gls{irr} considers the \emph{relative} similarity in ratings provided by multiple raters (i.e. the human subjects) over multiple targets (i.e. the term pairs), focusing on the order of the targets.
\gls{ira}, on the other hand, captures the \emph{absolute} homogeneity between the ratings, looking at the specific rating chosen by raters.

Several indices have been devised to capture \gls{irr} and \gls{ira} in psychological surveys \citep{banerjee:1999:beyond,lebreton:2008:answers}.
Most indices range between $0$ (total disagreement) and $1$ (perfect agreement).
For example, the ratings of two raters on three targets $\{1, 2, 3\}$ and $\{2, 3, 4\}$ obtain a $IRR=1$ and $IRA=0$: the subjects agree perfectly on the ordering of the targets, while disagreeing on all absolute ratings.
\citet{lebreton:2008:answers} recommend using several indices for \gls{irr} and \gls{ira}, to avoid the bias of any single index.   
We thus include the following indices of \gls{ira} and \gls{irr}: the mean Pearson's correlation coefficient \citep{rodgers:1988:thirteen}; Kendall's $W$ \citep{kendall:1939:problem}; Robinson's $A$ \citep{robinson:1957:statistical}; Finn coefficient \citep{finn:1970:note}; James, Demaree and Wolf's $r_{WG(J)}$ \citep{james:1984:estimating}.

The 305 responders included both native (208) and non-native English speakers (97).
We observed a substantially lower inter-subject agreement when including non-native speakers ($r_{WG(J)} < .5$):
the wider variability in these results is due to the varying knowledge of English of these subjects, who might have associated terms to `false friends' in their native language, i.e. expressions in two different languages that look or sound similar, but differ considerably in meaning.  
For example, Italian speakers may confuse the meaning of `factory' with `farm' (`fattoria' in Italian). 
Hence, they were excluded from the dataset.
Furthermore, three subjects did not complete the task, and their responses were discarded.
In order to detect random answers, we computed the correlation between every individual subject and the means.
This way, two subjects in the similarity test showed no correlation at all with the mean ratings (Spearman's $\rho \in [-.05,.05]$), and were removed from the dataset.

\begin{table}[t] 
\footnotesize
\begin{tabular}{lrccc} 
\hline
& & Relatedness & Similarity & Overall \\
& & $Q_{REL}$  & $Q_{SIM}$ & -- \\
\hline
\rowcolor{gray!15}
\textbf{Responders} & total $N$ & 81 & 122 & 203\\
\textbf{Gender} 
& Male   & 72 & 93 & 165 \\
& Female &  ~9 & 29 & ~38 \\
\rowcolor{gray!15}
\textbf{Age}
& 18-25 & 28 & 39 & ~67\\
\rowcolor{gray!15}
& 26-35 & 14 & 41 & ~55\\
\rowcolor{gray!15}
& 36-45 & 12 & 23 & ~35\\
\rowcolor{gray!15}
& 46-55 & 15 & 10 & ~25\\
\rowcolor{gray!15}
& 56-65 & ~7 & ~9 & ~16 \\
\rowcolor{gray!15}
& $>$65 & ~5 & - & ~~5 \\
\textbf{Continent} 
& Africa       & - & ~3& ~~3\\
& Asia         & - & ~1& ~~1\\
& Europe       & 58&95&  153\\
& North America& 11&20&  ~31\\
& South America& - & -& - \\
& Oceania      & 12& ~3& ~15\\
 \rowcolor{gray!15}
\textbf{Web map}
 & Never used     & ~6 & 14& ~20\\
 \rowcolor{gray!15}
\textbf{expertise} & Occasional user& 18 & 33& ~51\\
\rowcolor{gray!15}
 & Frequent user  & 37 & 39& ~76\\
 \rowcolor{gray!15}
& Expert         & 20 & 36& ~56\\
\hline
\end{tabular} 
\caption{Demographics of human subjects in \gls{oursurvey}}
\label{table:eval_demo}
\end{table}

Of the resulting dataset, Table \ref{table:eval_demo} summarises demographic information (age group, gender, continent of origin, and self-assessed map expertise).
As is possible to observe, the subjects tend be young, male, European, and frequent users of web maps.\footnote{Although a better gender, age, and geographic balances would be desirable, we found it difficult to obtain it in practice without drastically limiting the size of the sample.}
Table \ref{table:eval_ira} focuses on the indices of \gls{irr} and \gls{ira}.
Following \citet{resnik:1995:using}, we consider upper bound on the cognitive plausibility of a computable measure to be the highest correlation obtained by a human rater with the means (e.g. $\rho = .92$ for relatedness).
The table shows these upper bounds both for Spearman's $\rho$ and Kendall's $\tau$.
All the \gls{irr} and \gls{ira} indices indicate very similar results, falling in the range $[.61,.67]$.
Given the highly subjective nature of semantic conceptualisations, this correlation is satisfactory, and is comparable with analogous psychological surveys \citep{rodriguez:2004:comparing}.

Given the set of term pairs, and the set of human raters, we computed the relatedness/similarity scores as the \emph{mean ratings}, normalised in the interval $[0,1]$, where $0$ means no relatedness/similarity, and $1$ maximum relatedness/similarity.
As we have stated in the survey objectives, the distribution of such scores should be as even as possible, to ensure that a semantic measure performs well across the board, and not only in a specific region of the semantic relatedness/similarity space.
Several pairs in the dataset contain related but not similar terms, and the scores confirm this difference. 
More specifically, $<$\emph{sea,island}$>$ obtained a relatedness score of $.74$ and a similarity of $.4$.
Similarly, $<$\emph{mountain hut, mountaintop}$>$ obtained respectively $.71$ and $.49$ for relatedness and similarity.

\begin{table}[t] 
\footnotesize
\begin{tabular}{lrcc}
\hline
& & Relatedness & Similarity  \\
& & $Q_{REL}$  & $Q_{SIM}$  \\
\hline
 \rowcolor{gray!15}
\textbf{IRR}  & mean Pearson's $r$  & .64* & .65* \\
\textbf{IRA}  & Kendall's $W$ & .65* & .64*  \\
  & Robinson's $A$ & .62~ & .61~  \\
& Finn coefficient  & .65* & .66* \\
  & $r_{WG(J)}$  & .66~ & .67~ \\
 \rowcolor{gray!15}
\textbf{Upper bound}  & Spearman's $\rho$  & .92* & .93* \\
  \rowcolor{gray!15}
  & Kendall's $\tau$  & .79* & .82*  \\
\hline
 \end{tabular} 
\caption{Indices for interrater reliability (IRR) and interrater agreement (IRA) in \protect\gls{oursurvey}. (*) $p<.001$.}
\label{table:eval_ira}
\end{table}

A dimension that has not been addressed in existing similarity gold standards is that of the \emph{pair agreement}, i.e. the consistency of ratings expressed by all subjects on a single pair (see Section \ref{sec:relwork_goldStandards}).
For this purpose, we adopt James, Demaree and Wolf's $r_{WG}$, a popular index to measure \gls{ira} on a single target, based on the rating variance \citep{james:1984:estimating}.
Each pair in $Q_{REL}$ and $Q_{SIM}$ obtains an agreement measure $\in [0,1]$, where $0$ indicates a squared distribution (i.e. raters gave all ratings in equal proportion), and $1$ is perfect agreement (i.e. all raters assigned exactly the same rating to the pair).
Table \ref{table:geresid_pairs} shows the content of the resulting dataset, including mean ratings and pair agreement.

\begin{table}[t] 
\footnotesize
\begin{tabular}{rllcccc}
\hline
Pair  & &  & \multicolumn{2}{c}{Mean} & \multicolumn{2}{c}{Agreement}\\
\# & Term A & Term B & \sc{rel} & \sc{sim} & \sc{rel} & \sc{sim} \\
\hline
1 & motel & hotel & .93 & .90 & .86 & .82\\ 
2 & public transport station & railway platform & .87 & .81 & .80 & .72\\ 
3 & stadium & athletics track & .85 & .76 & .74 & .63\\ 
4 & theatre & cinema & .82 & .87 & .57 & .79\\ 
5 & art shop & art gallery & .78 & .75 & .58 & .60\\ 
 & \ldots & \ldots & \ldots & \ldots & \ldots & \ldots \\
46 & water ski facility & office furniture shop & .05 & .05 & .92 & .88\\ 
47 & greengrocer & aqueduct & .04 & .03 & .91 & .95\\ 
48 & interior decoration shop & tomb & .03 & .05 & .96 & .92\\ 
49 & political boundary & women's clothes shop & .02 & .02 & .96 & .93\\ 
50 & nursing home & continent & .01 & .02 & .97 & .95\\
\hline
 \end{tabular} 
\caption{Sample term pairs in \protect\gls{oursurvey}, with mean geo-semantic relatedness, similarity, and agreement}
\label{table:geresid_pairs}
\end{table}  
 
\begin{figure}[t]
  \centering 
  \includegraphics[width=43em]{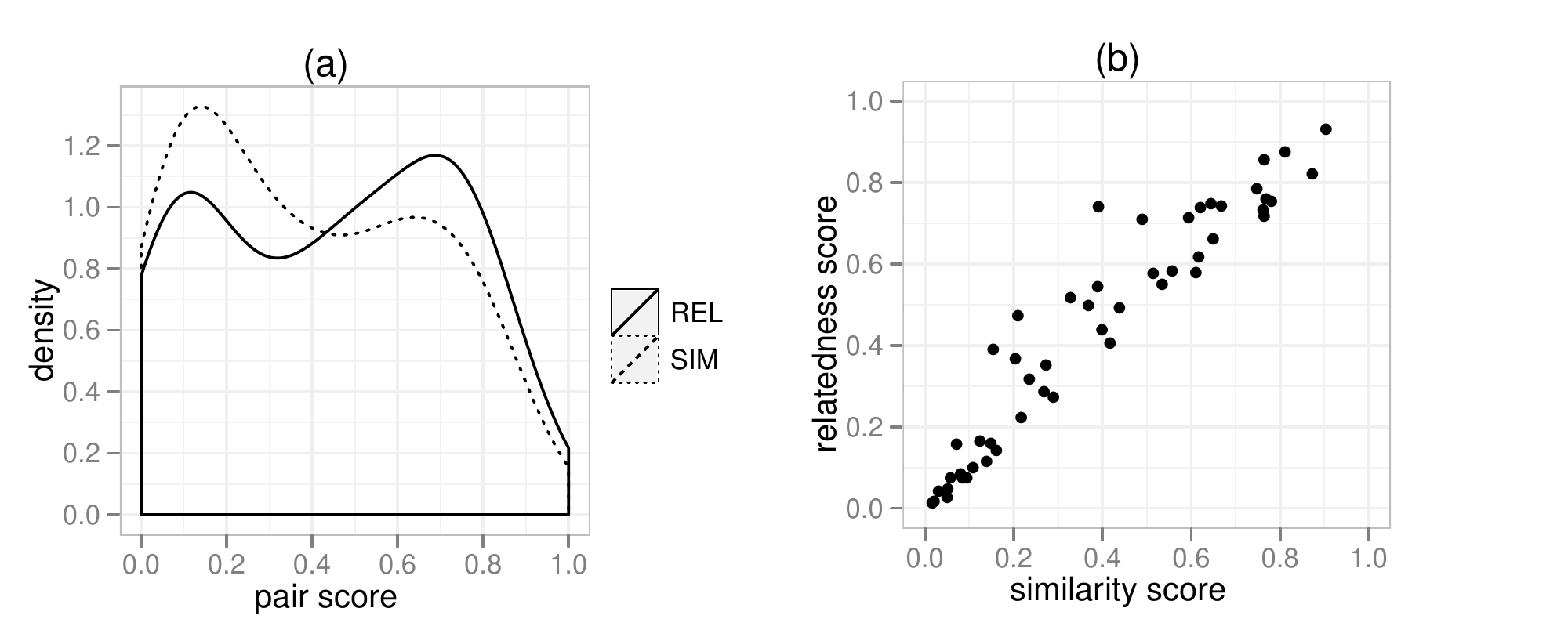}
  \caption{\protect\gls{oursurvey}: \textsc{rel}: semantic relatedness; \textsc{sim}: semantic similarity. (a) Density of pair score; (b) scatterplot of relatedness versus similarity.}
  \label{fig:relSimPlotDens}
\end{figure}

\begin{figure}[t]
  \centering 
  \includegraphics[width=41em]{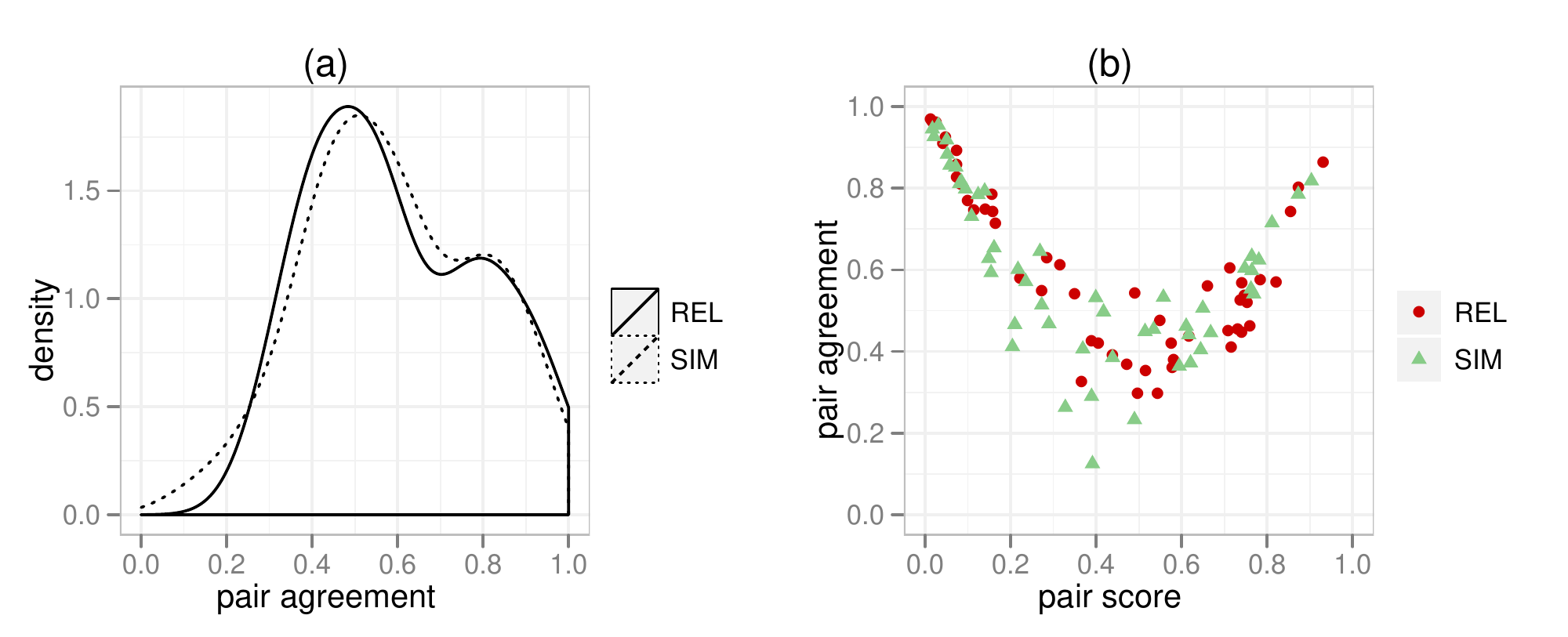}
  \caption{\protect\gls{oursurvey}: \textsc{rel}: semantic relatedness; \textsc{sim}: semantic similarity. (a) Density of pair agreement; (b) scatterplot of pair agreement and pair score.}
  \label{fig:relSimPlotAgr}
\end{figure}

Figures \ref{fig:relSimPlotDens} and \ref{fig:relSimPlotAgr} show several statistical characteristics of the resulting dataset, for the 50 pairs in $Q_{REL}$ and $Q_{SIM}$.
Plot \ref{fig:relSimPlotDens}(a) shows the density of the final relatedness/similarity scores, i.e. the normalised mean rankings in range $[0,1]$.
While the similarity is skewed towards the range $[0,.4]$, the relatedness has slightly more scores in the range $[.4,1]$, resulting in symmetrical densities.
This clearly reflects the fact that semantic similarity is a specific type of semantic relatedness, and semantic similarity is generally lower than relatedness.
This can be also observed in the sum of the 50 relatedness scores ($sum=22.01, mean=.44$) against the similarity scores ($sum=19.5, mean=.39$).
The paired Wilcoxon signed rank test \citep{wilcoxon:1945:individual} indicates that the relatedness scores are higher than the corresponding similarity ones, at $p < .001$.
This trend is clearly visible in plot \ref{fig:relSimPlotDens}(b).
Overall, these densities show that all the score range $[0,1]$ is satisfactorily covered, i.e. the dataset does not show large gaps.

Plots \ref{fig:relSimPlotAgr}(a) and \ref{fig:relSimPlotAgr}(b) show the properties of pair agreement (index $r_{WG}$), reporting the relationship between relatedness and similarity, the density of pair agreement, and the relationship between pair agreement and relatedness/similarity scores.
In terms of pair agreement, relatedness and similarity follow very close patterns, with a peak $\approx .5$.
This agreement might seem low, but it is largely expected, due to the subjective interpretation of the values on the Likert scale.

An explanation of this trend in the pair agreement lies in the fact that humans give consistently different ratings to the same objects: some subjects tend to be strict, and some lenient, resulting in different relative ratings, and therefore low absolute pair agreement \citep{lebreton:2008:answers}.
In this regard, a clear pattern emerges from plot \ref{fig:relSimPlotDens}(b).
Pair agreement tends to be high ($> .7$) at the extremes of the scores, when the relatedness/similarity judgement is very low ($[0,.25)$, no relation) or very high ($(.75,1]$, strong relation).
On the other hand, pairs with middle scores (in the interval $[.25,.75]$) tend to have low pair agreement.
Relatedness and similarity do not show important differences with respect to pair agreement ($sum=30, mean=.6$ for relatedness, $sum=29.6, mean=.59$ for similarity).
This detailed analysis, in particular in relation to \gls{irr} and \gls{ira}, confirms the statistical soundness of \gls{oursurvey}, which can be used to assess the cognitive plausibility of computational measures of \gls{grelsim}.

\section{Conclusions}
\label{sec:concl}

\glsreset{irr}
\glsreset{ira}
\glsreset{oursurvey}

To date, despite its great potential in \gls{gir} and information integration, geo-semantic relatedness has been only marginally studied.
In this article, we have discussed a notion of geo-semantic relatedness based on Lehrer's theory of semantic fields, contrasting it with the widely studied geo-semantic similarity.
Despite the variety and importance of computational measures devised in \gls{nlp}, the evaluation of such measures remains a difficult and complex task \cite{ferrara:2013:evalrelatedness}. 

In order to provide an evaluative baseline for geo-semantic research on relatedness and similarity, we have designed, collected, and analysed the \gls{oursurvey}. 
This dataset contains human judgements about 50 term pairs on semantic relatedness and similarity, covering 97 unique geographic terms.
To increase the dataset's usability and clarity, the terms have been mapped to the corresponding terms in \gls{wn}.
The judgements were collected from 203 English native speakers, through a randomised online survey.
\gls{oursurvey} is freely available online, released under an Open Knowledge license.\homepagenote{Datasets} 
The following points deserve particular consideration:
\begin{itemize}
  \item The human judgements have \gls{ira} and \gls{irr} in the interval $[.61,.67]$. Considering the type of psychological test, this is a fair agreement, indicating that the dataset can be used to evaluate computational measures of semantic relatedness and similarity for geographic terms.
  \item Human subjects strongly agree on cases of very high and low semantic relationships, and tend to have lower agreement on the intermediate cases.
  \item Semantic relatedness and similarity are strongly correlated ($\tau = .84, \rho = .95$). Furthermore, semantic relatedness scores are consistently higher than semantic similarity, confirming the more specific nature of semantic similarity.
  \item The contribution of \gls{oursurvey} lies both in its design and validation methodology, as well as the dataset itself. The raw data and the resulting dataset are available for analysis and re-use under a Creative Commons license.
  \item \gls{oursurvey} constitutes an evaluative baseline to evaluate measures of semantic similarity and relatedness.
  Furthermore, it permits the empirical determination of whether a given measure better approximates similarity or relatedness through the direct comparison of rankings or scores.
  \item A variety of techniques can be used to compare the rankings or scores generated by a computational measure with \gls{oursurvey}, including correlation coefficients (Spearman's $\rho$ and Kendall's $\tau$), and categorical approaches (Cohen's kappa or Fisher's exact test).
\end{itemize}

\noindent Although \gls{oursurvey} provides a novel resource to evaluate computational measures of \gls{grelsim}, several questions remain open.
\gls{oursurvey} distinguishes between \gls{grelsim}, but not among different \emph{contexts}.
As context has been identified as a key aspect of semantic similarity \cite{kessler:2007:similarity}, new datasets should be generated to capture explicitly the differences in geo-similarity and relatedness judgements with respect to different contexts, such as appearance and affordances.
The investigation of what specific geographic aspects are used by subjects in their judgements also constitutes important future work.
The dataset's \gls{ira} and \gls{irr} are comparable to similar datasets, but have a large margin of improvement.

As \citet{ferrara:2013:evalrelatedness} point out, this evaluative approach has several limitations.
Human subjects understand intuitively semantic relatedness and similarity, but the translation of such judgements into a number is very subjective.
Different information communities can express different judgements on the same term pairs.
Alternative approaches to the evaluation of computational measures should be investigated, aiming at cross-validating the findings generated by \gls{oursurvey}.
A promising route might consist of evaluating human-readable \emph{explanations} of relatedness measures, and not only numeric scores or rankings \cite{hecht:2012:explanatorysr}.
Moreover, the collection of judgements was conducted through online surveys in an uncontrolled environment, which have well-known issues \cite{Wright:2005:researching}.

Ultimately, the cognitive plausibility is assessed using correlation indexes such as Spearman's $\rho$ and Kendall's $\tau$, which have specific limitations.
For example, they tend to attribute the same weight to high and low similarity rankings, whilst computational applications normally need more precision on highly-related/similar pairs, which tend to be utilised in \gls{gir} and information integration.
Using \gls{oursurvey} as input data, new techniques to assess cognitive plausibility can be developed, offering tools tailored to the study of \gls{grelsim}.
Fruitful future work, as geo-semantic similarity is a specific case of geo-semantic relatedness, will consist of the generalisation of existing geo-similarity theories to the framework of geo-semantic relatedness.

\begin{acknowledgements}
The research presented in this article was funded by a Strategic Research Cluster grant (07/SRC/I1168) by Science Foundation Ireland under the National Development Plan. The authors gratefully acknowledge this support.
\end{acknowledgements}

\bibliographystyle{spbasicemph_fixed}      
\bibliography{thesis,mypub}

\end{document}